\pdfoutput=1

\documentclass[11pt]{article}
\usepackage[inkscapelatex=false]{svg}
\usepackage{enumitem}
\usepackage{acl2023}
\usepackage{tabularray}
\usepackage{times}
\usepackage{latexsym}
\usepackage{booktabs,tabularx}
\newcolumntype{Y}{>{\raggedright\arraybackslash}X} 
\usepackage[T1]{fontenc}

\usepackage[utf8]{inputenc}

\usepackage{microtype}
\usepackage[toc,page]{appendix}
\usepackage{inconsolata}
\usepackage{multirow}
\usepackage{graphicx}
\usepackage{tcolorbox} 
\usepackage{inconsolata}
\usepackage{booktabs}
\usepackage{dingbat}
\usepackage{rotating}
\usepackage{array,cellspace}

\definecolor{lightgray}{gray}{0.9}
\definecolor{lightgreen}{HTML}{6AA84F}
\definecolor{blue}{HTML}{4A86E8}
\definecolor{purple}{HTML}{9900FF}
\definecolor{red}{HTML}{FF0000}
\setlist[itemize,1]{leftmargin=\dimexpr 15pt}

\newtcolorbox{mybox}[3][]
{
  colframe = #2!25,
  colback  = #2!10,
  coltitle = #2!20!black,  
  title    = {#3},
  #1,
}

\title{Words Matter: Reducing Stigma in Online Conversations about Substance Use with Large Language Models}

\author{Layla Bouzoubaa, Elham Aghakhani, Rezvaneh Rezapour \\
Drexel University\\
  \texttt{\{lb3338, ea664, shadi.rezapour\}@drexel.edu} \\}

\begin{document}
\maketitle

\begin{abstract}
Stigma is a barrier to treatment for individuals struggling with substance use disorders (SUD), which leads to significantly lower treatment engagement rates. With only 7\% of those affected receiving any form of help, societal stigma not only discourages individuals with SUD from seeking help but isolates them, hindering their recovery journey and perpetuating a cycle of shame and self-doubt. This study investigates how stigma manifests on social media, particularly Reddit, where anonymity can exacerbate discriminatory behaviors. We analyzed over 1.2 million posts, identifying 3,207 that exhibited stigmatizing language towards people who use substances (PWUS). Using Informed and Stylized LLMs, we develop a model for de-stigmatization of these expressions into empathetic language, resulting in 1,649 reformed phrase pairs.
Our paper contributes to the field by proposing a computational framework for analyzing stigma and destigmatizing online content, and delving into the linguistic features that propagate stigma towards PWUS. Our work not only enhances understanding of stigma's manifestations online but also provides practical tools for fostering a more supportive digital environment for those affected by SUD. 
\end{abstract}
\section{Introduction}
\textbf{\textcolor{blue}{Warning:}} \textcolor{blue} {This paper includes language and content that may be offensive or triggering.} Every day, people struggling with substance use disorders (SUD) face a pervasive and often hidden enemy: stigma. This stigma, often deeply ingrained in societal attitudes, can act as a significant barrier to treatment and recovery. In fact, only approximately 7\% of people living with an SUD receive any form of treatment \cite{samhsa2023}, with stigma reported as a major barrier \cite{cdc2023}.  
SUD is a critical public health challenge in the US and worldwide, and the substantial stigma associated with these conditions only exacerbates the problem. 

Traditional support systems, although beneficial, often remain underutilized due to their perceived inaccessibility or the overwhelming stigma surrounding SUD, thus rendering this topic a societal taboo.

\begin{table}[t]
\centering
\small
\resizebox{0.9\columnwidth}{!}{%
\begin{tabular}{cp{0.95\columnwidth}} 
\toprule
\textbf{\texttt{Type}} & \textbf{\texttt{Statement}} \\
\midrule
\multirow{1}{*}{\rotatebox{90}{\texttt{Original}}} & \texttt{{\color[HTML]{FF0000} I have no empathy for drug addicts.} I had friends and family who have {\color[HTML]{FF0000} struggled with the ``disease''.} Everyone knows what happens {\color[HTML]{FF0000} when you start, and you usually end up dead.} Many of my old friends have become {\color[HTML]{FF0000}addicts} and I don't understand especially the ones with kids.} \\
\midrule
\multirow{1}{*}{\rotatebox{90}{\texttt{De-stigmatized}}} & \texttt{{\color[HTML]{0070C0} I find it difficult to empathize with individuals facing substance use challenges.} I had friends and family {\color[HTML]{0070C0} who encountered these difficulties.} It's widely acknowledged that {\color[HTML]{0070C0} there are risks involved from the outset, and the outcomes are often heartbreaking.} Several of my old friends have {\color[HTML]{0070C0} dealt with these challenges,} and it's particularly perplexing to me when they are parents.} \\
\bottomrule
\end{tabular}}
\caption{Example of directed stigmatizing language. De-stigmatized version generated with our Informed + Stylized model using GPT-4 removed stereotypes and harmful context while preserving the tone ({\color[HTML]{FF0000}stigma is in red}, {\color[HTML]{0070C0} destigmatized counterparts is in blue}).}
\label{tab:example}
\vspace{-0.5cm}
\end{table}
Social media platforms like Reddit have emerged as important spaces for community discussions \cite{bouzoubaa2023exploring, bouzoubaa-etal-2024-decoding}. However, the anonymity provided by these environments sometimes exacerbates stigmas, leading to discrimination. People suffering from SUD often encounter derogatory comments, judgment, or misinformation online \cite{schomerus2011stigma}, which can reinforce self-stigma and stop them from seeking help. The spread of stigmatizing attitudes on social media can also influence public opinion, further perpetuating the stereotypes and prejudices against those with SUD \cite{McLaren_2023}. As a result, despite the potential for support, the digital space can mirror and magnify the very societal stigmas it has the power to dismantle, affecting individuals' mental health and recovery processes adversely \cite{matsumoto2021perceived, mcneil2021understanding}.

The widespread stigma surrounding SUD requires urgent and innovative solutions. Leveraging technology and social media, we can develop empathetic, supportive interventions that fight against this stigma \cite{rahaman2023ai}. While research has explored mental health conversations and public perceptions on social media \cite{robinson2019measuring}, there remains a significant gap in efforts to destigmatize language in these discussions. Addressing this gap is crucial for fostering a more understanding and supportive environment for those affected by SUD.

Our work explores this opportunity and examines how stigmatizing language manifests in online communities and what solutions can be applied for de-stigmatizing such narratives (Table~\ref{tab:example}). Our study focuses on two research questions:

\begin{itemize}[noitemsep,topsep=0pt]
    \item[-] \textbf{RQ1:} How does stigmatizing language manifest in non-drug-related Reddit communities when discussing SUD, and what are the underlying factors that contribute to such expressions?
    \item[-] \textbf{RQ2: } How can we leverage LLMs to effectively de-stigmatize language, and what factors influence the success of this process?
\end{itemize}

To address these research questions, we collected over 1.2 million posts from non-drug-related subreddits, identifying 3,207 posts containing stigmatizing language towards people who use substances (PWUS). Leveraging large language models (LLMs), we developed a framework to characterize stigma based on conceptualization of~\citet{link_conceptualizing_2001} (\textit{labeling, stereotyping, separation, status loss, and discrimination}) and transform them into more empathetic versions, resulting in 1,649 de-stigmatized pairs. Our analysis showed that stimulants and cannabis were the most frequently mentioned substances, with stigma more generally being associated with interpersonal relationships and moral judgments. Human evaluations showed that our Informed + Stylized system using GPT-4 can reduce stigma while preserving the original tone and relevance. Automatic evaluations further confirmed that our approach effectively reduced stigma while maintaining the stylistic and psycholinguistic properties of the original posts.

Our work makes several key contributions: (1) public release of a unique dataset of labeled stigmatizing posts; (2) demonstration of frameworks for de-stigmatizing text; and (3) exploration of the linguistic characteristics of stigma expressions towards people who use substances (PWUS) online. Additionally, this study introduces innovative uses of LLMs for generating suggestions to mitigate potentially harmful language.
\section{Related Work}

\subsection{Stigma and Language}

Stigma, a complex social phenomenon, is deeply intertwined with language. The linguistic relativity principle, as described by \citet{Whorf1956}, suggests that language shapes our perception of reality, including the formation of stigmatizing views. In the context of substance use experiences (SUE) and SUD, stigma can manifest in multiple forms: \textit{self-stigma}, often rooted in shame \cite{luoma_slow_2012}; \textit{public stigma}, negative attitudes and beliefs which lead to discrimination and social exclusion; \textit{structural stigma}, which limits resources and opportunities, embedded in societal norms and institutional practices \cite{hatzenbuehler_structural_2016}.

Building upon \citet{goffman2009stigma}'s foundational work, \citet{link_conceptualizing_2001} conceptualized stigma as the co-occurrence of labeling, stereotyping, separation, status loss, and discrimination. This framework highlights how stigma operates alongside power inequalities, influencing both the individual and society at large.
Research has explored the manifestation of stigma in online communities \cite{nippert2021media}, particularly within social media platforms \cite{clark2021weight}, revealing both the potential for support and the amplification of existing stigmas, particularly among mental health and opiate-dedicated online communities~\cite{chen_examining_2022,eschliman_first-hand_2024}. 

Linguistic analysis has proven valuable in identifying and characterizing stigmatizing language. Dehumanizing labels and biased language can perpetuate negative stereotypes and contribute to discrimination \cite{giorgi2023lived}. A recent study by the CDC found that while stigmatizing language in traditional media has decreased over time, its use on social media platforms has increased \cite{McLaren_2023}, highlighting the need for targeted interventions in these spaces. The specific linguistic cues that distinguish stigmatizing content can differ between those with lived experience of substance use and those without, particularly regarding language considered ``othering'' and the use of labels like ``addict'' \cite{giorgi2023lived}.

\subsection{LLMs and Social Impact}
LLMs have shown promise in addressing social issues like hate speech detection \cite{Guo2023Investigation} and bias mitigation \cite{schlicht2024pitfalls}. Recent research demonstrates that LLMs can perform on par with or even surpass benchmark machine learning models in identifying hate speech \cite{kumarage2024harnessing}. Moreover, carefully crafted prompting strategies can leverage the knowledge encoded in LLMs to improve the detection of nuanced and context-dependent forms of hate speech \cite{guo-2023-investigation}. However, the application of LLMs in sensitive domains raises ethical concerns. The ``black box'' nature of these models can make it difficult to understand their decision-making processes, raising issues of transparency and accountability \cite{guo2024large}. Additionally, biases in training data can be inadvertently perpetuated, leading to discriminatory outcomes \cite{mei-2023-bias}. Addressing these ethical considerations is important for the responsible and equitable use of LLMs in de-stigmatization efforts.

\subsection{De-stigmatization Efforts}
Language-based interventions, such as the use of person-first language and empathetic communication, have shown promise in reducing stigma related to substance use. Research has demonstrated the impact of specific word choices on perceptions of individuals with SUD \cite{Kelly_Dow_Westerhoff_2010}. \citet{McGinty_2018} proposed a set of communication strategies to reduce stigma, including the use of sympathetic narratives, removing blame, and highlighting structural barriers to treatment. These findings contributed notably as the National Institute on Drug Abuse (NIDA) has also published guidelines for using non-stigmatizing language in discussions of SUD \cite{NIDA_WordsMatter}.

AI-mediated interventions, particularly those leveraging LLMs, have the potential to scale and automate de-stigmatization efforts. While prior work has focused on text detoxification and bias reduction, in general, \cite{dale-etal-2021-text, mendelsohn2020framework, pryzant2020automatically}, the specific application to SUD-related stigma remains underexplored. Additionally, \citet{spata2024substance} highlights the importance of using appropriate and well-validated measures to assess the effectiveness of interventions aimed at reducing stigma.

Our work builds upon the previous work by introducing a comprehensive computational approach to identify and categorize stigma. 
Focusing on public stigma, which we refer to as \textit{directed stigma} throughout the paper, we operationalize \citet{link_conceptualizing_2001}'s framework, analyzing instances of labeling, stereotyping, separation, and discrimination towards PWUS in discussions in non-drug-related Reddit communities.
\section{Data} \label{sec:data}

To achieve the study's objective of addressing stigmatizing language, we specifically focused on non-drug-related subreddits. This choice was made to capture how stigmatizing language manifests externally rather than within communities where members discuss their own experiences with drug use. Within these communities, stigmatizing language is often directed towards oneself (e.g., ``No one should hire a junkie like me, I'm useless'') or describes situations where members felt stigmatized (e.g., ``My co-workers stopped having lunch with me when they learned I've been to rehab twice'') which differs from the external stigmatizing language we aim to address. By focusing on non-drug-related subreddits, we ensure that our analysis targets the perpetuation of harmful stereotypes by those outside the drug-using community. This methodological choice is informed by the need to differentiate between internal and external stigma, as highlighted in the literature on stigma (e.g., \citet{link_conceptualizing_2001}'s attributes of stigma).

\paragraph{Data Collection. }
To investigate the manifestation of stigmatizing language in non-drug-related online communities, we collected data from four popular subreddits: \textit{r/unpopularopinion}, \textit{r/offmychest}, \textit{r/medicine}, and \textit{r/nursing}. The first two subreddits were chosen for their high activity levels, diverse user bases, and relevance to discussions of substance use and SUDs. Recent research has highlighted the prevalence of stigmatizing language within medical professional communities as well on platforms such as Twitter, although the overall use of stigmatizing and de-stigmatizing language was found to be low \cite{scott_graham_opioid_2022}. Given the critical role that healthcare professionals play in the lives of individuals with SUD, we included two of the most popular subreddits for healthcare professionals; \textit{r/nursing} and \textit{r/medicine}. 

We collected a total of 3.8 million posts from these subreddits. Table~\ref{tab:subreddit} shows the number of posts per subreddit. To ensure data quality, we excluded posts that were removed, deleted, or associated with deleted accounts. Additionally, we filtered out posts where the combined title and body text were less than 10 words to focus on substantive discussions. This resulted in a final dataset of 1.51 million posts for analysis.

\begin{table}[]
\centering
\resizebox{\columnwidth}{!}{%
\begin{tabular}{@{}llll@{}}
\toprule
\textbf{\texttt{Subreddit}} & \textbf{\texttt{\# Subscribers}} & \textbf{\texttt{\# Posts}} & \textbf{\texttt{Date Range}} \\
\cmidrule(r){1-1} 
\cmidrule(lr){2-2} 
\cmidrule(lr){3-3} 
\cmidrule(l){4-4} 
\texttt{r/medicine}         & 478K & 116,702   & 05/2005 - 12/2022 \\
\texttt{r/nursing}          & 715K & 212,755   & 12/2009 - 12/2022 \\
\texttt{r/offmychest}       & 3.2M & 1,607,341 & 02/2010 - 12/2022 \\
\texttt{r/unpopularopinion} & 4.3M & 2,044,463 & 08/2013 - 12/2022\\\bottomrule
\end{tabular}%
}
\caption{Selected subreddits and raw post and subscriber counts as of July 2024}
\label{tab:subreddit}
\vspace{-0.6cm}
\end{table}

\section{Methodology}
\begin{figure*}[t]
    \centering
    \includegraphics[width=0.95\textwidth]{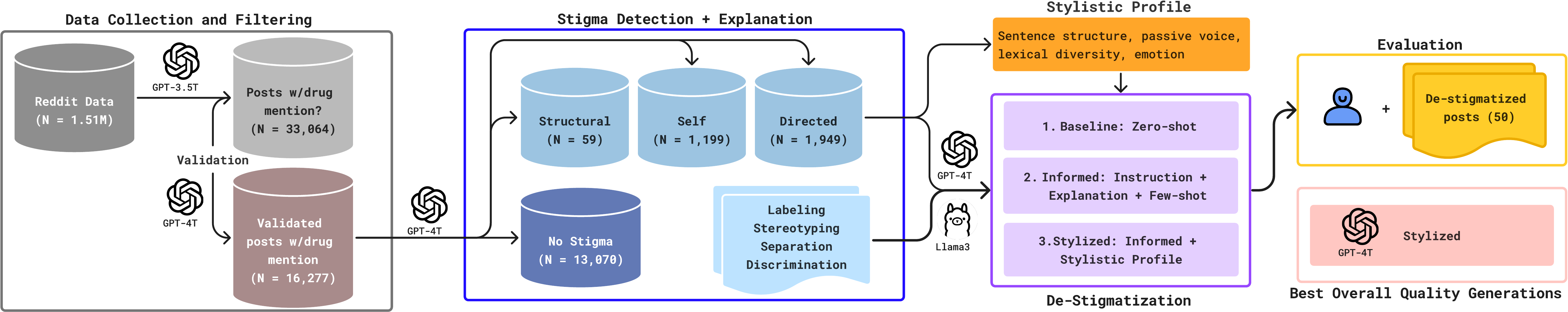}
    \caption{Full de-stigmatization pipeline.}
    \label{fig:pipeline}
\end{figure*}

To develop a stigma detection model and destigmatize texts, we first need to filter posts related to substance use. This is followed by detection and de-stigmatization processes. Figure~\ref{fig:pipeline} shows our study's overall pipeline. Each step is detailed in the following sections.

\subsection{Developing a Stigma Detection Model}
\subsubsection{Filtering Substance Use-Related Posts}
To identify posts containing stigmatizing language related to substance use, we first filtered posts collected from non-drug-related subreddits to find relevant discussions. Drug-related content includes any mention of illicit drugs or drug use (e.g., heroin, cocaine, LSD), prescription drugs that can be misused (e.g., narcotics, benzodiazepines), and other drugs that are not prescription but are also commonly misused (e.g., inhalants, bath salts). We began by manually annotating a random sample of 200 posts to establish a ground truth for relevance. Two annotators independently assessed each post, achieving 100\% agreement on the presence or absence of substance use-related content. 

Given the nuanced nature of language around substance use, including slang and idiomatic expressions, we used LLMs with few-shot prompting to identify posts within the larger dataset. 
Based on a comprehensive assessment of performance metrics, including precision, recall, F1-score, and estimate time (see Appendix \ref{app:time}), we selected GPT-3.5 Turbo as the most suitable model for this task. As a result of Task 1, we identified around 33,064 posts containing at least one mention of drugs or drug-related content.

\paragraph{Validation Layer. } Given the tendency of GPT-3.5 to overgeneralize, we implemented a validation layer using GPT-4 Turbo to re-evaluate all posts initially flagged as containing substance use-related content (N = 33,064).
To evaluate the effectiveness of this validation layer, we randomly sampled 725 posts from the GPT-3.5 output (252 labeled as drug-related (\textit{D}) and 473 as non-drug-related (\textit{ND})) and conducted a manual evaluation. The posts labeled as \textit{D} by GPT-3.5 were then passed through the GPT-4 validation layer. Out of the 252 posts initially labeled as \textit{D}, 212 were confirmed as \textit{D} by GPT-4, resulting in an $F1 = 0.86$. From the 33,064 posts labeled as \textit{D} by GPT-3.5, 16,277 were validated as \textit{D} by GPT-4.

\subsubsection{Extracting Stigmatizing Language}
The posts labeled as containing drug content were then labeled for their inclusion of stigmatizing language. Stigmatizing language could be in the form of directed language towards PWUS that perpetuates harmful stereotypes, expressions of internalized stigma (i.e., self-stigma), or illustrations of structural or systemic stigma (e.g., criminal justice towards PWUS in the United States). To do this, we took a random sample of 200 posts from the 16,277 posts labeled \textit{D} and manually annotated for the inclusion of stigmatizing language. Any posts that contained directed stigmatizing language were also broken down into four attributes: 1) labeling, 2) stereotyping, 3) loss of power, and 4) discrimination. This process was re-iterated several times until substantial agreement was met ($k= 0.67$). The remaining posts were then labeled using GPT-4 Turbo using the prompt in Appendix \ref{app:prompts}.  

\paragraph{Explainability of Stigma Detection. }
In the pursuit of transparency and interpretability, we incorporated an explanation layer into our stigma detection model. 
Specifically, when the model identified a post as containing directed stigma towards PWUS, it was prompted to provide a detailed explanation for its classification by identifying the specific instances within the text that corresponded to each of the four elements of stigma outlined by \citet{link_conceptualizing_2001}: labeling, stereotyping, separation, and discrimination, mimicking our annotation process.

\begin{table}[]
\centering
\resizebox{\columnwidth}{!}{%
\begin{tabular}{@{}lcccc@{}}
\toprule
                       & \multicolumn{4}{c}{\texttt{\textbf{Stigma Type}}} \\ \midrule
\texttt{\textbf{Substance Category}} & \texttt{\textbf{Directed}} & \texttt{\textbf{Self}} & \texttt{\textbf{Structural}} & \texttt{\textbf{Total}} \\
\texttt{Stimulants}             & 818      & 380      & 20      & 1218     \\
\texttt{Cannabis}               & 515      & 276      & 27      & 818      \\
\texttt{Narcotics}              & 501      & 250      & 18      & 769      \\
\texttt{Depressants}            & 92       & 102      & 6       & 200      \\
\texttt{Hallucinogens}          & 90       & 68       & 4       & 162      \\
\texttt{Reversal Agents}        & 38       & 3        & 0       & 41       \\
\texttt{Drugs of Concern}       & 7        & 7        & 0       & 14       \\
\texttt{Synthetic Cannabinoids} & 11       & 3        & 0       & 14       \\
\texttt{Other}                  & 4        & 3        & 1       & 8        \\
\texttt{Designer Drugs}         & 6        & 0        & 0       & 6        \\
\texttt{Unspecified}            & 537      & 475      & 9       & 1021     \\
\bottomrule
\end{tabular}%
}
\caption{Cross-tabulation of substance categories mentioned in a post by the type of stigmatizing language used. Note that multiple substance categories may be mentioned in the same post.}
\label{tab:stigma_substance}
\vspace{-0.8cm}
\end{table}

\subsection{De-Stigmatizing Problematic Language}
To address and mitigate the impact of stigmatizing language in texts, we used two different LLMs across three different Models. Our objective is to determine which model is most effective at transforming stigmatizing language into expressions that are more empathetic and inclusive.

\noindent\textbf{Model 1: Baseline. }
In the baseline phase, we explored the capabilities of two LLMs in zero-shot de-stigmatization: GPT-4 Turbo and Llama 3-70B-Instruct. We provided the models with the original stigmatizing post and instructed them to generate a de-stigmatized version without any additional context or guidance. This approach allowed us to assess the inherent de-stigmatization capabilities of these models in the absence of explicit knowledge or stylistic refinements.

\noindent\textbf{Model 2: Informed LLM. }
Inspired by the principles of ``Constitutional AI,'' \cite{Bai_2022} we enhanced the LLM prompts in Phase 2 with explicit instructions, definitions, and explanations related to stigma. Constitutional AI refers to the development and operation of AI models that adhere to the principles and legal standards, ensuring respect for human rights, ethical guidelines, and public accountability. 
Drawing upon the insights gained from our analysis of stigmatizing language (RQ1), we provided the model with a structured understanding of the four stigma elements (labeling, stereotyping, separation, and discrimination) and their manifestations in the context of substance use.

\begin{itemize}[noitemsep,topsep=0pt]
    \item[-] \textbf{Labeling}: The model was instructed to identify and reword any labeling instances in the post, guided by a definition, explanation, and examples from RQ1 analysis.
    \item[-] \textbf{Stereotyping, Separation, and Discrimination:} The model was tasked with addressing these three interrelated elements of stigma simultaneously. The prompt included definitions for each element, examples from RQ1 analysis, and an explanation as to why these elements are harmful to guide the LLM to mitigate these forms of stigma through rephrasing, reframing, or adding context.
\end{itemize}

\noindent By incorporating these explicit instructions and structured explanation of stigma, we aimed to guide the LLM in generating de-stigmatized outputs that actively addressed each of the four stigma elements identified in the original post.

\noindent\textbf{Model 3: Informed LLM + Stylistic Considerations. }
Building upon the informed LLM approach of Phase 2, we further refined the de-stigmatization process by incorporating stylistic considerations. We aimed to ensure that the de-stigmatized output not only addressed the harmful content but also maintained the original post's emotional tone and stylistic features. To achieve this, we employed a combination of techniques:
\begin{itemize}[noitemsep,topsep=0pt]
    \item[-] \textbf{Emotion Analysis:} We used a pre-trained, RoBERTa \cite{Liu_2019} model fine-tuned on the GoEmotions dataset \cite{demszky_goemotions_2020} \footnote{\url{https://huggingface.co/SamLowe/roberta-base-go_emotions}}, to classify the emotional tone of the original post and instructed the LLM to preserve this tone in the de-stigmatized version.
    \item[-] \textbf{Punctuation and Syntax:} We analyzed the use of punctuation and sentence structure (i.e. sentence length variation) in the original post and encouraged the LLM to replicate these patterns in the output.
    \item[-] \textbf{Stylistic Elements:} Posts were analyzed for phrase style, specifically the measure of textual lexical diversity (MTLD) \cite{mccarthy2010mtld} and the use of passive voice, to ensure that the de-stigmatized output maintained the original post's overall writing style.
\end{itemize}

These elements, plus the explanations, were used to produce de-stigmatized outputs that were less harmful and stylistically congruent with the original post, thereby maintaining the author's voice and reducing the potential for inauthenticity.

\subsubsection{Evaluation of De-Stigmatized Posts}
 
\noindent\textbf{Human Evaluation. } To assess the effectiveness of our six systems (baseline, informed, and informed + stylized for GPT-4 and Llama3), we conducted a human evaluation with five reviewers on a random sample of 110 posts (a total of 660 generated texts). Our reviewers come from a variety of backgrounds, including HCI, NLP, and Social Computing. To evaluate the systems, we instructed our reviewers to analyze the generated text from each model and rank the models based on the overall quality, the extent of de-stigmatization, and the faithfulness of the outputs. Following traditional NLG assessments, quality was evaluated on criteria including naturalness, cohesion, human-likeness, and overall coherence \cite{howcroft2020twenty}. The assessment of de-stigmatization was judged based on removing 
negative or harmful stereotypes, and the systems with the least amount of labeling, stereotyping, separation, status loss, and discrimination. Faithfulness was evaluated based on the amount of transferred information from the original post without unnecessary details \cite{sai2022survey}. Comprehensive evaluation guideline is provided in Appendix~\ref{app: human eval}.

\paragraph{Automatic Evaluation. }To further evaluate the stylistic similarity between original posts and their de-stigmatized counterparts generated by our models, we conducted a linguistic analysis using LIWC \cite{boyd2022development}. We then performed a t-test to compare the linguistic features identified in both the original and de-stigmatized texts. Given the unique nature of our task, traditional metrics such as BLEU \cite{papineni-etal-2002-bleu} or ROUGE \cite{lin-2004-rouge} were deemed unsuitable because the generated text and its original counterparts differ significantly in meaning. Additionally, the absence of pre-existing de-stigmatized versions of these texts prevented us from conducting comparative analyses with an established benchmark.

\section{Experimental Results \& Analysis}  \label{sec:results}
\subsection{Characteristics of Stigmatizing Language}
\paragraph{Mentioned Substances. } Out of 16,277 posts discussing drugs, our stigma detection pipeline resulted in 3,207 posts containing stigmatizing language (Figure~\ref{fig:pipeline}). Of these, 1,949 posts contained directed stigma, 59 represented systemic/structural stigma and 1,199 contained self-stigmatizing language. 
As shown in Table~\ref{tab:stigma_substance}, analysis of stigmatizing posts revealed that stimulants like ``meth'' (methamphetamine) and ``coke'' (cocaine) were the most frequently mentioned drug categories, followed by cannabis (``weed'', ``pot'') for all types of stigma. Posts that mentioned drug use terms like ``drugs'', ``high'', or ``pills,'' but no specific substance were categorized as ``Unspecified.''

\noindent\textbf{Anatomy of Stigma. } To further understand \textit{who}, did \textit{what}, and \textit{why} in the context of stigma towards PWUS in online discussions, we examined representative entities, subject-verb pairs, and topic models. Representative entities and subject-verb pairs reveal the \textit{direction} of the mentions (who), while entity and substance frequencies highlight the targets of stigma (what). Topic modeling allows us to infer the underlying motivations and contexts of stigmatizing language (why). For this purpose, we used a multifaceted linguistic analysis: we first extracted subject-verb pairs using part of speech tagging in \textit{spaCy} \cite{spacy}, classified emotions toward these pairs in each post using GoEmotions \cite{demszky_goemotions_2020} and RoBERTa \cite{Liu_2019}, and performed topic modeling with BERTopic \cite{grootendorst2022bertopic} and KeyBERT \cite{grootendorst2020keybert}. 

Within the posts showing directed stigma (Appendix \ref{app:data}), we primarily observe expressions of \textit{sadness} and \textit{annoyance}, with some \textit{neutrality}. Notably, interpersonal relationships surface as a key theme, featuring mentions of family members like ``sister,'' ``dad,'' and ``mother'' alongside substances like ``cannabis'' and ``amphetamines.'' This aligns well with the overall prevalence of stimulants and cannabis in substance mentions (Table \ref{tab:stigma_substance}). The dominant topic, ``Cannabis and Legalization Stigma'' centers on these substances, often referred to as ``it,'' in a \textit{neutral} tone primarily related to ``smoking.'' Following closely is ``Stigma Toward Interpersonal Relationships,'' characterized by expressions of knowledge (\textit{I know}) from the subject \textit{``I''} directed towards family members, often tinged with \textit{sadness}. Another notable topic, ``Moral Judgments of Others,'' reveals \textit{annoyance} (\textit{I hate}) towards individuals like ``neighbors,'' ``homeless,'' and ``junkies'' associated with ``heroin'' and other drugs. 

Shifting to self-stigmatizing posts, we find distinct emotional undertones and actions. While interpersonal entities are less prominent compared to directed stigma, these posts feature more action verbs and a wider variety of substances. The primary topic, ``Depression around Sobriety,'' is marked by expressions of possession (\textit{I have}) and state of being (\textit{I am}) in relation to \textit{depression}, \textit{sobriety}, and \textit{quitting}. Disturbingly, another topic reveals a \textit{desire} for \textit{overdose}, specifically mentioning narcotics like ``fentanyl,'' ``dilaudid,'' and ``tramadol,'' alongside benzodiazepines like ``Xanax'' and ``clonazepam,'' a combination known to be potentially lethal due to respiratory arrest.

Finally, while only a few posts showed structural stigma (N = 59), making it hard to analyze topics, the emotions in these posts were mainly neutral.

\subsection{De-Stigmatization with LLMs}

\begin{table*}[t]
\centering
\resizebox{0.7\textwidth}{!}{%
\begin{tabular}{@{}lcccc@{}}
\toprule
\textbf{\texttt{Model}}& \textbf{\texttt{LLM}} & \textbf{\texttt{Best Overall Quality}} & \textbf{\texttt{Most De-Stigmatized}} & \textbf{\texttt{Most Faithful}} \\ \midrule
\textbf{\texttt{Informed + Stylized}} & \texttt{GPT4}  & \textbf{37}                   & 18                           & \textbf{49}            \\
\textbf{\texttt{Informed}} & \texttt{GPT4}             & 24                            & 7                            & 33                     \\
\textbf{\texttt{Informed}} & \texttt{Llama}            & 19                            & 8                            & 16                     \\
\textbf{\texttt{Informed + Stylized}}& \texttt{Llama} & 13                            & 3                            & 6                      \\
\textbf{\texttt{Baseline}} & \texttt{Llama}            & 9                             & 32                           & 2                      \\
\textbf{\texttt{Baseline}} & \texttt{GPT4}             & 6                             & \textbf{40}                  & 2                      \\ \bottomrule
\end{tabular}%
}
\caption{Frequency of evaluation metrics by systems for 110 de-stigmatized posts.}
\label{tab:eval-results}
\vspace{-0.7cm}
\end{table*}


\noindent\textbf{Human Evaluation.} Human evaluation of 110 randomly selected and destigmatized posts is shown in Table \ref{tab:eval-results}. 
Our reviewers found the Informed + Stylized system with GPT-4 the best ``Overall Quality'' for de-stigmatizing posts and the ``Most Faithful'', striking a balance between reducing stigma and maintaining relevance and tone with the original message. Interestingly, the ``Most De-stigmatized'' system was the Baseline with GPT-4. While this system may be effective at removing overt stigmatizing language, the lack of contextual awareness and stylistic preservation resulted in posts sounding generic, insincere, unfaithful, or disconnected from the original poster's voice. For example, with the Baseline + GPT-4, this post:

\emph{``Decriminalizing drugs is a dangerous and dumb idea and shouldn't be done. what a wonderful idea, let's give druggies more room and less consequences for 
doing the hard stuff. 
when out, people need to keep an eye out to not accidentally step on used needles, crime is 
up from junkies who 
do anything to afford their next fix.''}
is de-stigmatized to: 

\emph{``Decriminalizing drugs is a complex and controversial topic that merits a thoughtful discussion. There are valid concerns about the potential increase in visibility and accessibility of substances.
states that have adopted more lenient policies towards drug possession, 
see increased public drug use and 
challenges, such as discarded needles and 
crime, which 
worries local residents.''}

For practical applications, this could affect the model's ability to foster genuine empathy and understanding in online communities. Table \ref{tab:example} displays a successful de-stigmatized post using our best model. The revised post is less harmful and stereotypical but maintains the speaker's tone and overall message.

\noindent\textbf{Automatic Evaluation. } We conducted a stylistic similarity analysis using LIWC to compare original stigmatizing posts with their de-stigmatized versions generated by our top-rated system (Informed + Stylized GPT-4). A pairwise two-way t-test showed no significant differences in means across all LIWC variables between the two sets of posts. While certain categories like bigwords (use of six-letter words or more) and cogproc (cognitive processes) were more common in de-stigmatized posts, the overall psycholinguistic properties were largely maintained. This result is promising as it shows our de-stigmatization approach effectively reduced stigma while preserving the original style and emotional tone, essential for authenticity.

\vspace{-0.25cm}
\section{Discussion}
\noindent\textbf{Stigma also stems from personal connections. } Our findings showed a complex landscape of stigma within non-drug-related online communities where discussions about substance use often become entangled with interpersonal relationships and ingrained societal biases - particularly towards specific substances, namely stimulants (e.g., methamphetamine) and cannabis (e.g., ``weed,'' ``pot''). The frequent mentions of these substances within a stigmatizing context may reflect societal concerns about their visibility and impact, aligning with our topic modeling results, where the dominant topic in directed stigma is ``Cannabis Legalization Stigma.'' These findings highlight the role of close relationships (family, friends) in both expressing and experiencing stigma. For instance, within the topic ``Interpersonal Stigma,'' we observe individuals expressing sadness and using the verb ``know'' when discussing family members struggling with substance use. This underscores the need for de-stigmatization efforts to extend beyond public forums and into private spheres, as stigma from close social circles can be particularly harmful due to the emotional weight and potential for isolation \cite{luoma_slow_2012}.

The online nature of these interactions presents a duality of stigma manifestations that is important to understand when developing any intervention. While anonymity might offer a shield for individuals to express stigmatizing views they might suppress offline, it could also create a space for open dialogue and support. The disinhibition afforded by online platforms could lead to more candid discussions about SUD, potentially challenging stigma through shared experiences and mutual understanding. However, it may also create a space for misinformed judgments and harmful stereotypes, as anonymity can reduce accountability.

When considering de-stigmatization efforts, any digital intervention should consider the social actors in addition to the social constructs (e.g. hospitals, employers). This would be considerably important in collectivist communities (e.g. Indian or Middle Eastern) where stigma towards family members with an SUD (i.e. \textit{affiliate stigma}) may prevent families from providing the necessary medical support to their loved ones and ultimately delaying treatment \cite{corrigan2006blame}. 

\paragraph{LLMs can be guided by explanation and stylistic information. }
In our de-stigmatization efforts, we intentionally avoided providing the LLMs with a rigid definition of ``de-stigmatized.'' Instead, we adopted a more nuanced approach, drawing inspiration from the principles of ``Constitutional AI'' \cite{Bai_2022} and prior work on text detoxification and bias reduction using LLMs \cite{dale-etal-2021-skoltechnlp, mendelsohn-etal-2021-modeling, pryzant2020automatically}. We focused on explaining why specific phrases might be problematic and instructed the model to address these issues, constitutionally, while preserving the original style. 
For instance, to tackle separation, the LLMs were guided to draw equivalences between individuals with SUD and those without, emphasizing shared humanity. Labeling was addressed by replacing derogatory terms like ``junkie'' with person-centered language like ``person with a substance use disorder,'' mitigating the over-generalization tendencies of LLMs. Stereotyping and discrimination were handled by re-framing generalizations and removing any implications of discrimination, promoting a more empathetic understanding of individuals struggling with SUD.
 
\noindent\textbf{Most de-stigmatized does not mean most pragmatic. }
While the baseline model removes stigmatizing language, it often does so at the expense of nuance and context. For instance, evaluators noted that the baseline model sometimes ``terribly misunderstood the post,'' resulting in generic or insincere responses that failed to capture the original poster's intent. This highlights the importance of removing stigma and preserving the authenticity and emotional tone of the original message.
Our findings emphasize the importance of striking a balance between promoting empathetic language and providing overly refined language, which might trivialize the experiences of individuals with SUD or avoid addressing the root causes of stigma.

\vspace{-0.25cm}
\section{Conclusion}
This study investigated the manifestations of stigma towards PWUS in four popular non-drug-related subreddits (\textit{r/unpopularopinion}, \textit{r/offmychest}, \textit{r/nursing}, \textit{r/medicine}). We identified 3,207 posts containing one of three main types of stigma (self, structural, and directed). Given the contextual nuance of self and structural stigma, we focused our efforts on de-stigmatizing instances of directed stigma (N = 1,649). Experimenting with three different models and two different LLMs (GPT-4 and Llama), the model that used the conceptualization of stigma \cite{link_conceptualizing_2001}, few-shot examples, and the original post's stylistic profile generated the most faithful and appropriate destigmatized texts. Our exploration of LLM-based de-stigmatization demonstrates the potential of these models to transform harmful language into more empathetic expressions while emphasizing the importance of preserving authenticity and the original poster's voice. 
While our focus has been on SUD stigma, the insights and methodologies presented here have broader implications for understanding and addressing stigma related to other marginalized groups. Future work could explore the role of misinformation in perpetuating stigma and leverage external knowledge bases (e.g. DrugBank) to develop more informed and effective de-stigmatization strategies. By integrating these approaches, we can create a more supportive and inclusive online environment for individuals affected by stigma, ultimately promoting understanding, empathy, and recovery.

\newpage
\section{Limitations}
Our findings primarily apply to English-speaking populations on one specific social media platform, which may not be generalizable to other linguistic or cultural contexts. We selected certain subreddits based on our assessment of relevance, which may have limited the breadth of our data; exploring additional subreddits could potentially provide a more comprehensive view. The performance and accuracy of the models we used, dependent on their training data, may not capture all nuances of stigmatizing language. Despite our ethical considerations, the automated analysis of sensitive topics like SUD carries risks of misinterpretation, necessitating ongoing research and continuous evaluation of ethical challenges in using large language models.

\section{Ethics Statement}
We acknowledge the diversity of perspectives on substance use and advocate for harm reduction strategies. All data was publicly available at the time of collection, and no direct interaction occurred between researchers and users. Our research was exempt from review by our institution's Internal Review Board (IRB). We adhere to strict data protection measures and have slightly altered any quotes to preserve anonymity and post integrity.  Our goal is not to erase personal experiences but to reframe them in less harmful ways, aligned with the original sentiment. The discussions in this paper should not be interpreted to suggest anyone's lived experience is more valid than another.

\bibliography{references, anthology,layla-zotero}

\begin{thebibliography}{49}
\expandafter\ifx\csname natexlab\endcsname\relax\def\natexlab#1{#1}\fi

\bibitem[{Bai et~al.(2022)Bai, Kadavath, Kundu, Askell, Kernion, Jones, Chen, Goldie, Mirhoseini, McKinnon, Chen, Olsson, Olah, Hernandez, Drain, Ganguli, Li, Tran-Johnson, Perez, Kerr, Mueller, Ladish, Landau, Ndousse, Lukosuite, Lovitt, Sellitto, Elhage, Schiefer, Mercado, DasSarma, Lasenby, Larson, Ringer, Johnston, Kravec, Showk, Fort, Lanham, Telleen-Lawton, Conerly, Henighan, Hume, Bowman, Hatfield-Dodds, Mann, Amodei, Joseph, McCandlish, Brown, and Kaplan}]{Bai_2022}
Yuntao Bai, Saurav Kadavath, Sandipan Kundu, Amanda Askell, Jackson Kernion, Andy Jones, Anna Chen, Anna Goldie, Azalia Mirhoseini, Cameron McKinnon, Carol Chen, Catherine Olsson, Christopher Olah, Danny Hernandez, Dawn Drain, Deep Ganguli, Dustin Li, Eli Tran-Johnson, Ethan Perez, Jamie Kerr, Jared Mueller, Jeffrey Ladish, Joshua Landau, Kamal Ndousse, Kamile Lukosuite, Liane Lovitt, Michael Sellitto, Nelson Elhage, Nicholas Schiefer, Noemi Mercado, Nova DasSarma, Robert Lasenby, Robin Larson, Sam Ringer, Scott Johnston, Shauna Kravec, Sheer~El Showk, Stanislav Fort, Tamera Lanham, Timothy Telleen-Lawton, Tom Conerly, Tom Henighan, Tristan Hume, Samuel~R. Bowman, Zac Hatfield-Dodds, Ben Mann, Dario Amodei, Nicholas Joseph, Sam McCandlish, Tom Brown, and Jared Kaplan. 2022.
\newblock \href {https://doi.org/10.48550/arXiv.2212.08073} {Constitutional ai: Harmlessness from ai feedback}.
\newblock (arXiv:2212.08073).
\newblock ArXiv:2212.08073 [cs].

\bibitem[{Bouzoubaa et~al.(2024)Bouzoubaa, Aghakhani, Song, Trinh, and Rezapour}]{bouzoubaa-etal-2024-decoding}
Layla Bouzoubaa, Elham Aghakhani, Max Song, Quang Trinh, and Shadi Rezapour. 2024.
\newblock \href {https://aclanthology.org/2024.findings-acl.367} {Decoding the narratives: Analyzing personal drug experiences shared on {R}eddit}.
\newblock In \emph{Findings of the Association for Computational Linguistics ACL 2024}, pages 6131--6148, Bangkok, Thailand and virtual meeting. Association for Computational Linguistics.

\bibitem[{Bouzoubaa et~al.(2023)Bouzoubaa, Young, and Rezapour}]{bouzoubaa2023exploring}
Layla Bouzoubaa, Jordyn Young, and Rezvaneh Rezapour. 2023.
\newblock Exploring the landscape of drug communities on reddit: A network study.
\newblock In \emph{Proceedings of the International Conference on Advances in Social Networks Analysis and Mining}, pages 558--565.

\bibitem[{Boyd et~al.(2022)Boyd, Ashokkumar, Seraj, and Pennebaker}]{boyd2022development}
Ryan~L Boyd, Ashwini Ashokkumar, Sarah Seraj, and James~W Pennebaker. 2022.
\newblock The development and psychometric properties of liwc-22.
\newblock \emph{Austin, TX: University of Texas at Austin}, pages 1--47.

\bibitem[{{Centers for Disease Control and Prevention}(2023)}]{cdc2023}
{Centers for Disease Control and Prevention}. 2023.
\newblock Reducing stigma to prevent opioid overdose.
\newblock \url{https://www.cdc.gov/stop-overdose/stigma-reduction/index.html}.
\newblock Accessed: 2024-06-15.

\bibitem[{Chen et~al.(2022)Chen, Johnny, and Conway}]{chen_examining_2022}
Annie~T. Chen, Shana Johnny, and Mike Conway. 2022.
\newblock \href {https://doi.org/10.1016/j.dadr.2022.100061} {Examining stigma relating to substance use and contextual factors in social media discussions}.
\newblock \emph{Drug and Alcohol Dependence Reports}, 3:100061.

\bibitem[{Clark et~al.(2021)Clark, Lee, Jingree, O'Dwyer, Yue, Marrero, Tamez, Bhupathiraju, and Mattei}]{clark2021weight}
Olivia Clark, Matthew~M Lee, Muksha~Luxmi Jingree, Erin O'Dwyer, Yiyang Yue, Abrania Marrero, Martha Tamez, Shilpa~N Bhupathiraju, and Josiemer Mattei. 2021.
\newblock Weight stigma and social media: evidence and public health solutions.
\newblock \emph{Frontiers in nutrition}, 8:739056.

\bibitem[{Corrigan et~al.(2006)Corrigan, Watson, and Miller}]{corrigan2006blame}
Patrick~W Corrigan, Amy~C Watson, and Frederick~E Miller. 2006.
\newblock Blame, shame, and contamination: the impact of mental illness and drug dependence stigma on family members.
\newblock \emph{Journal of family psychology}, 20(2):239.

\bibitem[{Dale et~al.(2021{\natexlab{a}})Dale, Markov, Logacheva, Kozlova, Semenov, and Panchenko}]{dale-etal-2021-skoltechnlp}
David Dale, Igor Markov, Varvara Logacheva, Olga Kozlova, Nikita Semenov, and Alexander Panchenko. 2021{\natexlab{a}}.
\newblock \href {https://doi.org/10.18653/v1/2021.semeval-1.126} {{S}koltech{NLP} at {S}em{E}val-2021 task 5: Leveraging sentence-level pre-training for toxic span detection}.
\newblock In \emph{Proceedings of the 15th International Workshop on Semantic Evaluation (SemEval-2021)}, pages 927--934, Online. Association for Computational Linguistics.

\bibitem[{Dale et~al.(2021{\natexlab{b}})Dale, Voronov, Dementieva, Logacheva, Kozlova, Semenov, and Panchenko}]{dale-etal-2021-text}
David Dale, Anton Voronov, Daryna Dementieva, Varvara Logacheva, Olga Kozlova, Nikita Semenov, and Alexander Panchenko. 2021{\natexlab{b}}.
\newblock \href {https://doi.org/10.18653/v1/2021.emnlp-main.629} {Text detoxification using large pre-trained neural models}.
\newblock In \emph{Proceedings of the 2021 Conference on Empirical Methods in Natural Language Processing}, pages 7979--7996, Online and Punta Cana, Dominican Republic. Association for Computational Linguistics.

\bibitem[{Demszky et~al.(2020)Demszky, Movshovitz-Attias, Ko, Cowen, Nemade, and Ravi}]{demszky_goemotions_2020}
Dorottya Demszky, Dana Movshovitz-Attias, Jeongwoo Ko, Alan Cowen, Gaurav Nemade, and Sujith Ravi. 2020.
\newblock \href {http://arxiv.org/abs/2005.00547} {{GoEmotions}: {A} {Dataset} of {Fine}-{Grained} {Emotions}}.
\newblock ArXiv:2005.00547 [cs].

\bibitem[{Eschliman et~al.(2024)Eschliman, Choe, DeLucia, Addison, Jackson, Murray, German, Genberg, and Kaufman}]{eschliman_first-hand_2024}
E.L. Eschliman, K.~Choe, A.~DeLucia, E.~Addison, V.W. Jackson, S.M. Murray, D.~German, B.L. Genberg, and M.R. Kaufman. 2024.
\newblock \href {https://doi.org/10.1016/j.socscimed.2024.116772} {First-hand accounts of structural stigma toward people who use opioids on {Reddit}}.
\newblock \emph{Social Science and Medicine}, 347.

\bibitem[{Giorgi et~al.(2023)Giorgi, Bellew, Habib, Sherman, Sedoc, Smitterberg, Devoto, Himelein-Wachowiak, and Curtis}]{giorgi2023lived}
Salvatore Giorgi, Douglas Bellew, Daniel Roy~Sadek Habib, Garrick Sherman, Jo{\~a}o Sedoc, Chase Smitterberg, Amanda Devoto, McKenzie Himelein-Wachowiak, and Brenda Curtis. 2023.
\newblock Lived experience matters: automatic detection of stigma on social media toward people who use substances.
\newblock \emph{arXiv preprint arXiv:2302.02064}.

\bibitem[{Goffman(2009)}]{goffman2009stigma}
Erving Goffman. 2009.
\newblock \emph{Stigma: Notes on the management of spoiled identity}.
\newblock Simon and schuster.

\bibitem[{Grootendorst(2020)}]{grootendorst2020keybert}
Maarten Grootendorst. 2020.
\newblock Keybert: Minimal keyword extraction with bert.

\bibitem[{Grootendorst(2022)}]{grootendorst2022bertopic}
Maarten Grootendorst. 2022.
\newblock Bertopic: Neural topic modeling with a class-based tf-idf procedure.
\newblock \emph{arXiv preprint arXiv:2203.05794}.

\bibitem[{Guo et~al.(2023{\natexlab{a}})Guo, Hu, Mu, Shi, Zhao, Vishwamitra, and Hu}]{Guo2023Investigation}
Keyan Guo, Alexander Hu, Jaden Mu, Ziheng Shi, Ziming Zhao, Nishant Vishwamitra, and Hongxin Hu. 2023{\natexlab{a}}.
\newblock \href {https://doi.org/10.1109/ICMLA58977.2023.00237} {An investigation of large language models for real-world hate speech detection}.
\newblock In \emph{2023 International Conference on Machine Learning and Applications (ICMLA)}, pages 1568--1573.

\bibitem[{Guo et~al.(2023{\natexlab{b}})Guo, Hu, Mu, Shi, Zhao, Vishwamitra, and Hu}]{guo-2023-investigation}
Keyan Guo, Alexander Hu, Jaden Mu, Ziheng Shi, Ziming Zhao, Nishant Vishwamitra, and Hongxin Hu. 2023{\natexlab{b}}.
\newblock \href {https://doi.org/10.1109/ICMLA58977.2023.00237} {An investigation of large language models for real-world hate speech detection}.
\newblock In \emph{2023 International Conference on Machine Learning and Applications (ICMLA)}, pages 1568--1573.

\bibitem[{Guo et~al.(2024)Guo, Lai, Thygesen, Farrington, Keen, and Li}]{guo2024large}
Zhijun Guo, Alvina Lai, Johan~Hilge Thygesen, Joseph Farrington, Thomas Keen, and Kezhi Li. 2024.
\newblock Large language model for mental health: A systematic review.
\newblock \emph{arXiv preprint arXiv:2403.15401}.

\bibitem[{Hatzenbuehler(2016)}]{hatzenbuehler_structural_2016}
Mark~L. Hatzenbuehler. 2016.
\newblock \href {https://doi.org/10.1037/amp0000068} {Structural stigma: {Research} evidence and implications for psychological science}.
\newblock \emph{American Psychologist}, 71(8):742--751.
\newblock Place: US Publisher: American Psychological Association.

\bibitem[{Honnibal and Montani(2020)}]{spacy}
Matthew Honnibal and Ines Montani. 2020.
\newblock \href {https://spacy.io} {spacy: Industrial-strength natural language processing in python}.

\bibitem[{Howcroft et~al.(2020)Howcroft, Belz, Clinciu, Gkatzia, Hasan, Mahamood, Mille, Van~Miltenburg, Santhanam, and Rieser}]{howcroft2020twenty}
David~M Howcroft, Anya Belz, Miruna Clinciu, Dimitra Gkatzia, Sadid~A Hasan, Saad Mahamood, Simon Mille, Emiel Van~Miltenburg, Sashank Santhanam, and Verena Rieser. 2020.
\newblock Twenty years of confusion in human evaluation: Nlg needs evaluation sheets and standardised definitions.
\newblock In \emph{13th International Conference on Natural Language Generation 2020}, pages 169--182. Association for Computational Linguistics.

\bibitem[{Kelly et~al.(2010)Kelly, Dow, and Westerhoff}]{Kelly_Dow_Westerhoff_2010}
John~F. Kelly, Sarah~J. Dow, and Cara Westerhoff. 2010.
\newblock \href {https://doi.org/10.1177/002204261004000403} {Does our choice of substance-related terms influence perceptions of treatment need? an empirical investigation with two commonly used terms}.
\newblock \emph{Journal of Drug Issues}, 40(4):805–818.

\bibitem[{Kumarage et~al.(2024)Kumarage, Bhattacharjee, and Garland}]{kumarage2024harnessing}
Tharindu Kumarage, Amrita Bhattacharjee, and Joshua Garland. 2024.
\newblock Harnessing artificial intelligence to combat online hate: Exploring the challenges and opportunities of large language models in hate speech detection.
\newblock \emph{arXiv preprint arXiv:2403.08035}.

\bibitem[{Lin(2004)}]{lin-2004-rouge}
Chin-Yew Lin. 2004.
\newblock \href {https://aclanthology.org/W04-1013} {{ROUGE}: A package for automatic evaluation of summaries}.
\newblock In \emph{Text Summarization Branches Out}, pages 74--81, Barcelona, Spain. Association for Computational Linguistics.

\bibitem[{Link and Phelan(2001)}]{link_conceptualizing_2001}
Bruce~G. Link and Jo~C. Phelan. 2001.
\newblock \href {https://doi.org/10.1146/annurev.soc.27.1.363} {Conceptualizing {Stigma}}.
\newblock \emph{Annual Review of Sociology}, 27(1):363--385.
\newblock \_eprint: https://doi.org/10.1146/annurev.soc.27.1.363.

\bibitem[{Liu et~al.(2019)Liu, Ott, Goyal, Du, Joshi, Chen, Levy, Lewis, Zettlemoyer, and Stoyanov}]{Liu_2019}
Yinhan Liu, Myle Ott, Naman Goyal, Jingfei Du, Mandar Joshi, Danqi Chen, Omer Levy, Mike Lewis, Luke Zettlemoyer, and Veselin Stoyanov. 2019.
\newblock \href {https://doi.org/10.48550/arXiv.1907.11692} {Roberta: A robustly optimized bert pretraining approach}.
\newblock (arXiv:1907.11692).
\newblock ArXiv:1907.11692 [cs].

\bibitem[{Luoma et~al.(2012)Luoma, Kohlenberg, Hayes, and Fletcher}]{luoma_slow_2012}
Jason~B. Luoma, Barbara~S. Kohlenberg, Steven~C. Hayes, and Lindsay Fletcher. 2012.
\newblock \href {https://doi.org/10.1037/a0026070} {Slow and steady wins the race: {A} randomized clinical trial of acceptance and commitment therapy targeting shame in substance use disorders}.
\newblock \emph{Journal of Consulting and Clinical Psychology}, 80(1):43--53.
\newblock Publisher: American Psychological Association.

\bibitem[{Matsumoto et~al.(2021)Matsumoto, Santelices, and Lincoln}]{matsumoto2021perceived}
Atsushi Matsumoto, Claudia Santelices, and Alisa~K Lincoln. 2021.
\newblock Perceived stigma, discrimination and mental health among women in publicly funded substance abuse treatment.
\newblock \emph{Stigma and Health}, 6(2):151.

\bibitem[{McCarthy and Jarvis(2010)}]{mccarthy2010mtld}
Philip~M McCarthy and Scott Jarvis. 2010.
\newblock Mtld, vocd-d, and hd-d: A validation study of sophisticated approaches to lexical diversity assessment.
\newblock \emph{Behavior research methods}, 42(2):381--392.

\bibitem[{McGinty et~al.(2018)McGinty, Pescosolido, Kennedy-Hendricks, and Barry}]{McGinty_2018}
Emma McGinty, Bernice Pescosolido, Alene Kennedy-Hendricks, and Colleen~L. Barry. 2018.
\newblock \href {https://doi.org/10.1176/appi.ps.201700076} {Communication strategies to counter stigma and improve mental illness and substance use disorder policy}.
\newblock \emph{Psychiatric Services}, 69(2):136–146.

\bibitem[{McLaren et~al.(2023)McLaren, Jones, Noonan, Idaikkadar, and Sumner}]{McLaren_2023}
Nilay McLaren, Christopher~M. Jones, Rita Noonan, Nimi Idaikkadar, and Steven~A. Sumner. 2023.
\newblock \href {https://doi.org/10.1016/j.drugalcdep.2023.109807} {Trends in stigmatizing language about addiction: A longitudinal analysis of multiple public communication channels}.
\newblock \emph{Drug and Alcohol Dependence}, 245:109807.

\bibitem[{McNeil(2021)}]{mcneil2021understanding}
Sandra~R McNeil. 2021.
\newblock Understanding substance use stigma.
\newblock \emph{Journal of Social Work Practice in the Addictions}, 21(1):83--96.

\bibitem[{Mei et~al.(2023)Mei, Fereidooni, and Caliskan}]{mei-2023-bias}
Katelyn Mei, Sonia Fereidooni, and Aylin Caliskan. 2023.
\newblock \href {https://doi.org/10.1145/3593013.3594109} {Bias against 93 stigmatized groups in masked language models and downstream sentiment classification tasks}.
\newblock In \emph{Proceedings of the 2023 ACM Conference on Fairness, Accountability, and Transparency}, FAccT '23, page 1699–1710, New York, NY, USA. Association for Computing Machinery.

\bibitem[{Mendelsohn et~al.(2021)Mendelsohn, Budak, and Jurgens}]{mendelsohn-etal-2021-modeling}
Julia Mendelsohn, Ceren Budak, and David Jurgens. 2021.
\newblock \href {https://doi.org/10.18653/v1/2021.naacl-main.179} {Modeling framing in immigration discourse on social media}.
\newblock In \emph{Proceedings of the 2021 Conference of the North American Chapter of the Association for Computational Linguistics: Human Language Technologies}, pages 2219--2263, Online. Association for Computational Linguistics.

\bibitem[{Mendelsohn et~al.(2020)Mendelsohn, Tsvetkov, and Jurafsky}]{mendelsohn2020framework}
Julia Mendelsohn, Yulia Tsvetkov, and Dan Jurafsky. 2020.
\newblock A framework for the computational linguistic analysis of dehumanization.
\newblock \emph{Frontiers in artificial intelligence}, 3:55.

\bibitem[{{NIDA}(2023)}]{NIDA_WordsMatter}
{NIDA}. 2023.
\newblock Words matter - terms to use and avoid when talking about addiction.
\newblock {https://nida.nih.gov/research-topics/addiction-science/words-matter-preferred-language-talking-about-addiction}.
\newblock Accessed: 2024.

\bibitem[{Nippert et~al.(2021)Nippert, Tomiyama, Smieszek, and Incollingo~Rodriguez}]{nippert2021media}
Kathryn~E Nippert, A~Janet Tomiyama, Stephanie~M Smieszek, and Angela~C Incollingo~Rodriguez. 2021.
\newblock The media as a source of weight stigma for pregnant and postpartum women.
\newblock \emph{Obesity}, 29(1):226--232.

\bibitem[{Papineni et~al.(2002)Papineni, Roukos, Ward, and Zhu}]{papineni-etal-2002-bleu}
Kishore Papineni, Salim Roukos, Todd Ward, and Wei-Jing Zhu. 2002.
\newblock \href {https://doi.org/10.3115/1073083.1073135} {{B}leu: a method for automatic evaluation of machine translation}.
\newblock In \emph{Proceedings of the 40th Annual Meeting of the Association for Computational Linguistics}, pages 311--318, Philadelphia, Pennsylvania, USA. Association for Computational Linguistics.

\bibitem[{Pryzant et~al.(2020)Pryzant, Martinez, Dass, Kurohashi, Jurafsky, and Yang}]{pryzant2020automatically}
Reid Pryzant, Richard~Diehl Martinez, Nathan Dass, Sadao Kurohashi, Dan Jurafsky, and Diyi Yang. 2020.
\newblock Automatically neutralizing subjective bias in text.
\newblock In \emph{Proceedings of the aaai conference on artificial intelligence}, volume~34, pages 480--489.

\bibitem[{Rahaman et~al.(2023)Rahaman, Ahsan, Anjum, Rahman, and Rahman}]{rahaman2023ai}
Md~Saidur Rahaman, MM~Ahsan, Nishath Anjum, Md~Mizanur Rahman, and Md~Nafizur Rahman. 2023.
\newblock The ai race is on! google's bard and openai's chatgpt head to head: an opinion article.
\newblock \emph{Mizanur and Rahman, Md Nafizur, The AI Race is on}.

\bibitem[{Robinson et~al.(2019)Robinson, Turk, Jilka, and Cella}]{robinson2019measuring}
Patrick Robinson, Daniel Turk, Sagar Jilka, and Matteo Cella. 2019.
\newblock Measuring attitudes towards mental health using social media: investigating stigma and trivialisation.
\newblock \emph{Social psychiatry and psychiatric epidemiology}, 54:51--58.

\bibitem[{Sai et~al.(2022)Sai, Mohankumar, and Khapra}]{sai2022survey}
Ananya~B Sai, Akash~Kumar Mohankumar, and Mitesh~M Khapra. 2022.
\newblock A survey of evaluation metrics used for nlg systems.
\newblock \emph{ACM Computing Surveys (CSUR)}, 55(2):1--39.

\bibitem[{Schlicht et~al.(2024)Schlicht, Altiok, Taouk, and Flek}]{schlicht2024pitfalls}
Ipek~Baris Schlicht, Defne Altiok, Maryanne Taouk, and Lucie Flek. 2024.
\newblock Pitfalls of conversational llms on news debiasing.
\newblock \emph{arXiv preprint arXiv:2404.06488}.

\bibitem[{Schomerus et~al.(2011)Schomerus, Lucht, Holzinger, Matschinger, Carta, and Angermeyer}]{schomerus2011stigma}
Georg Schomerus, Michael Lucht, Anita Holzinger, Herbert Matschinger, Mauro~G Carta, and Matthias~C Angermeyer. 2011.
\newblock The stigma of alcohol dependence compared with other mental disorders: a review of population studies.
\newblock \emph{Alcohol and alcoholism}, 46(2):105--112.

\bibitem[{Scott~Graham et~al.(2022)Scott~Graham, Conway, Bottner, and Claborn}]{scott_graham_opioid_2022}
S.~Scott~Graham, F.N. Conway, R.~Bottner, and K.~Claborn. 2022.
\newblock \href {https://doi.org/10.1080/16066359.2022.2061962} {Opioid use stigmatization and destigmatization in health professional social media}.
\newblock \emph{Addiction Research and Theory}.

\bibitem[{Spata et~al.(2024)Spata, Gupta, Lear, Lunze, and Luoma}]{spata2024substance}
Angelica Spata, Ishita Gupta, M~Kati Lear, Karsten Lunze, and Jason~B Luoma. 2024.
\newblock Substance use stigma: A systematic review of measures and their psychometric properties.
\newblock \emph{Drug and Alcohol Dependence Reports}, page 100237.

\bibitem[{{Substance Abuse and Mental Health Services Administration}(2023)}]{samhsa2023}
{Substance Abuse and Mental Health Services Administration}. 2023.
\newblock Samhsa announces nsduh results detailing mental illness and substance use levels in 2021.
\newblock {https://www.samhsa.gov/newsroom/press-announcements/20230104/samhsa-announces-nsduh-results-detailing-mental-illness-substance-use-levels-2021}.

\bibitem[{Whorf(1956)}]{Whorf1956}
Benjamin~Lee Whorf. 1956.
\newblock \emph{Language, Thought, and Reality: Selected Writings of Benjamin Lee Whorf}.
\newblock MIT Press.

\end{thebibliography}
\bibliographystyle{acl_natbib}

\appendix
\section{Comparison of LLMs for Labeling Drug Mention} \label{app:time}
We examined various LLMs (combination of open-source and proprietary) to differentiate between drug-related and non-drug-related posts on Reddit, using a dataset of 200 manually annotated posts. To assess the performance of each model, we calculated the F-1 score, which is a measure of a test's accuracy that considers both precision and recall. Additionally, we analyzed the total time and cost required to process this amount of posts. These findings are detailed in the table provided in Table \ref{tab:task1_justify}. This table helps to illustrate not only the effectiveness of each model in terms of accuracy but also their efficiency and economic viability for similar tasks.

\begin{table}[h]
\centering
\resizebox{\columnwidth}{!}{%
\begin{tabular}{@{}lllll@{}}
\toprule
\textbf{\texttt{Model}}  & \textbf{F1} & \textbf{\texttt{Total Time}} & \textbf{\texttt{Cost (USD)}} & \textbf{\texttt{RPM}} \\
\cmidrule(r){1-1} 
\cmidrule(lr){2-2}
\cmidrule(lr){3-3} 
\cmidrule(lr){4-4} 
\cmidrule(l){5-5}
\texttt{GPT 3.5-Turbo} & 0.78        & 9.52 s              & 0.07                & 3,500*       \\
\texttt{GPT 4-Turbo}   & 0.9         & 19.05 s             & 1.31                & 500*         \\
\texttt{Mistral}       & 0.48        & 330.60 s            & 0                   & 300**        \\
\texttt{Llama3-8B}     & 0.38        & 59.9 s              & 0                   & 600***      \\ \bottomrule             
\end{tabular}%
}
\caption{Comparison on four LLMs considered to label 1.51M posts for the mention of drugs or drug use based on a random sample of 200 manually-annotated posts.\\
`*' based on OpenAI Tier 3 usage (see \url{https://platform.openai.com/docs/guides/rate-limits/usage-tiers?context=tier-three})\\
`**' based on Hugging Face Inference API rate limit per hour \\
`***' based on Together.ai API rate per second for Paid Tier (\url{https://docs.together.ai/docs/rate-limits}).
}

\label{tab:task1_justify}
\end{table}

\section{Prompts}\label{app:prompts}

\begin{figure*}[t]
    \centering
    \includegraphics[width=0.8\textwidth]{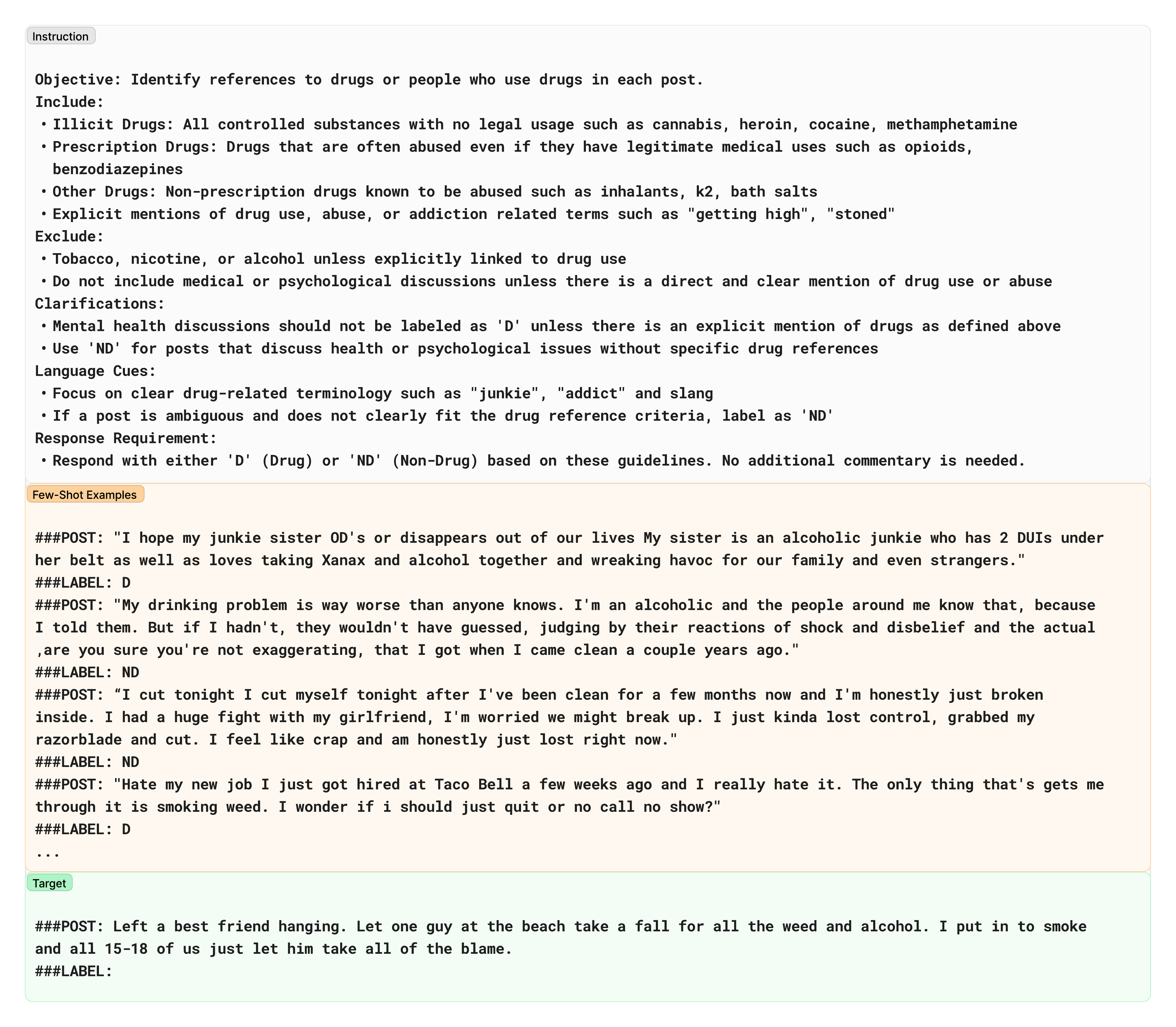}
    \caption{Few-shot prompting to determine whether a post contains a reference to illicit substance use.}
    \label{fig:task-1}
\end{figure*}

\begin{figure*}[t]
    \centering
    \includegraphics[width=0.8\textwidth]{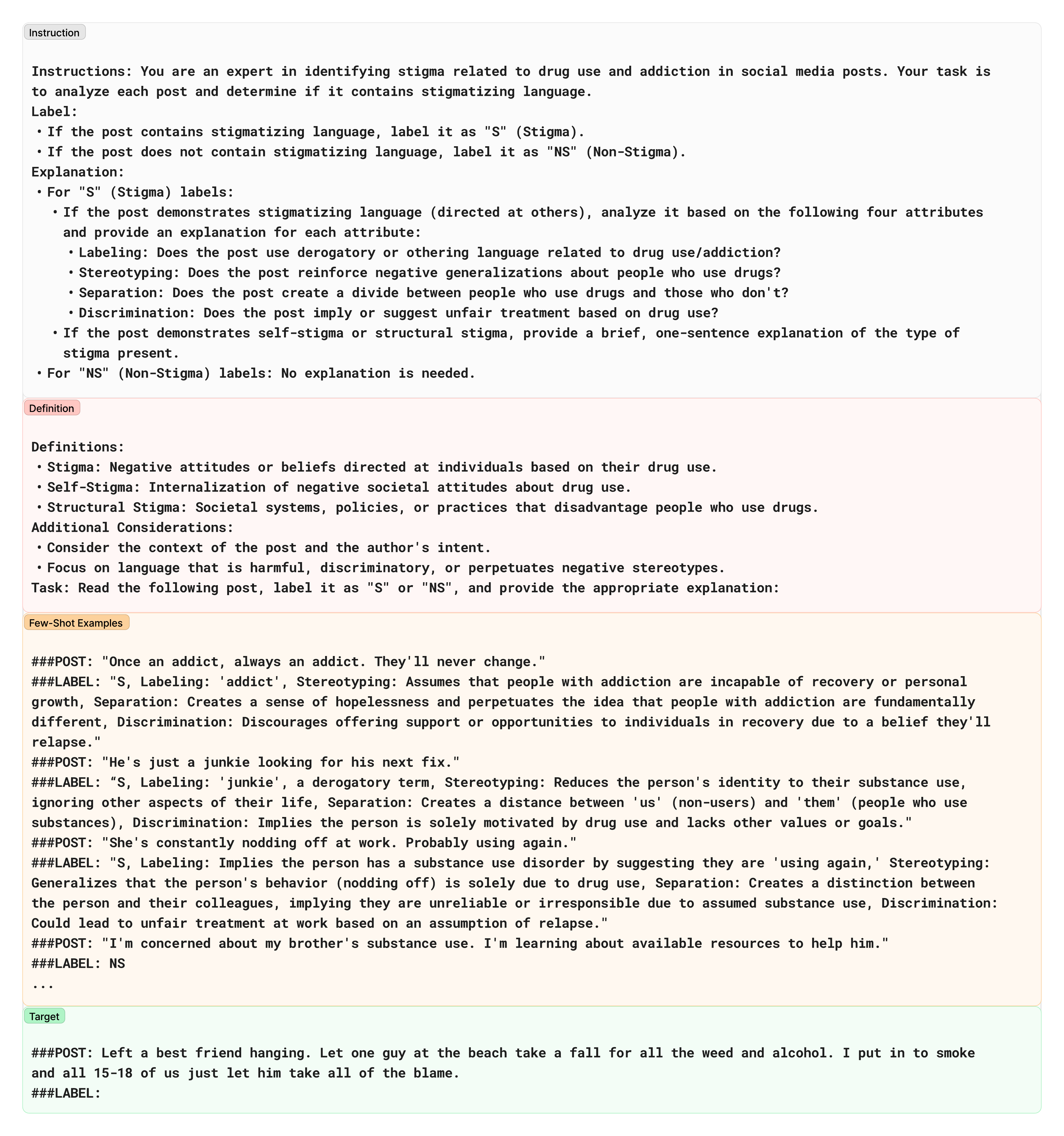}
    \caption{Few-shot prompting to determine whether a post contains stigmatizing language towards PWUS.}
    \label{fig:task-2}
\end{figure*}

\begin{figure*}[t]
    \centering
    \includegraphics[width=0.8\textwidth]{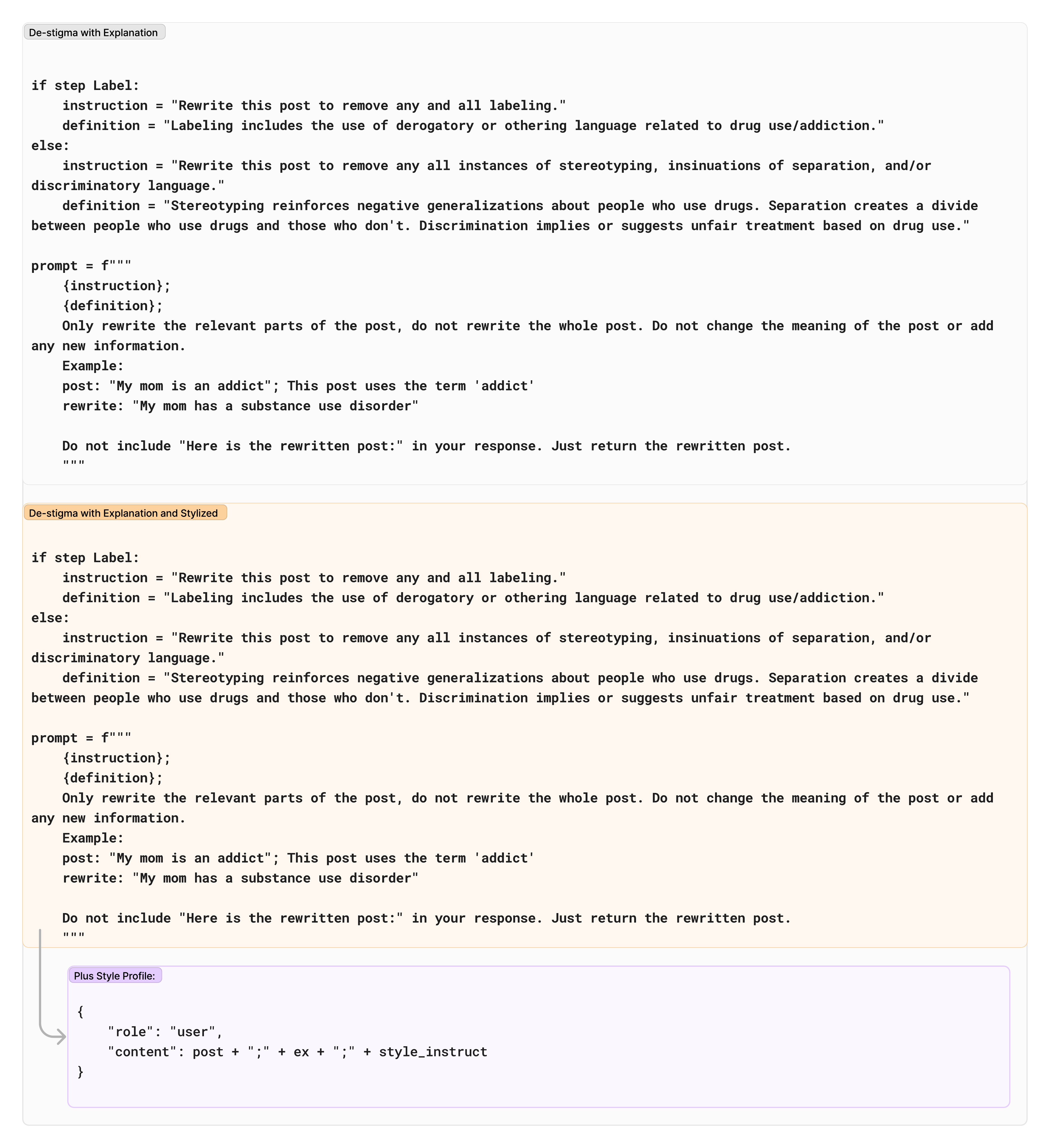}
    \caption{Few-shot prompting for de-stigmatizing language towards PWUS, explanation and explanation plus style profile.}
    \label{fig:task-3}
\end{figure*}

In our study, we implemented a multi-step pipeline using different prompts for each stage, which includes data filtering, stigma detection with explanations, and destigmatization. The prompts tailored for data filtering, stigma detection, and destigmatization are detailed in Figures \ref{fig:task-1}, \ref{fig:task-2} and \ref{fig:task-3}. This structured approach ensures efficient handling and analysis of stigmatizing content in social media posts.

\section{Data Analysis}\label{app:data}
In our study, we conducted a comprehensive linguistic analysis of online posts about drug use and addiction-related stigmas. We extracted and analyzed representative entities, subject-verb pairs, and utilized topic modeling to identify themes in direct and self-stigmatizing posts. These topics were organized by names, representative keywords, dominant emotions, and frequent verb-subject pairs, presented in Tables \ref{tab:summary1} and \ref{tab:summary2}. For linguistic processing, we used \textit{spaCy} for subject-verb extraction, GoEmotions and RoBERTa for emotion classification, and BERTopic and KeyBERT for topic modeling. This multi-dimensional approach provided a detailed understanding of the discourse within these posts.

\begin{table*}[t]
\centering
\small
\resizebox{\textwidth}{!}{
\begin{tabular}{>{\raggedright\arraybackslash}p{2cm}>{\raggedright\arraybackslash}p{3.2cm}>{\raggedright\arraybackslash}p{2cm}>{\raggedright\arraybackslash}p{3cm}>{\raggedright\arraybackslash}p{8.5cm}}
\toprule
\textbf{\texttt{Name}} & \textbf{\texttt{Representation}} & \textbf{\texttt{Top Emotion}} & \textbf{\texttt{Top Verb-Subject Pairs}} & \textbf{\texttt{Example}} \\
\midrule
\texttt{Cannabis Legalization Stigma} & \texttt{\textbf{\textcolor{lightgreen}{marijuana, cannabis, weed,}} drugs, \textbf{\textcolor{lightgreen}{addicts,}} \textbf{\textcolor{purple}{sober, smoking,}} \textbf{\textcolor{lightgreen}{heroin, pot,}} \textbf{\textcolor{blue}{smokers}}} & \texttt{neutral} & \texttt{\{`it', `is'\}: 126, \{`i', `have'\}: 117, \{`i', `know'\}: 91} & \texttt{Your addiction and dependence isn't slighter than mines and vice versa. Just because weed doesn't have physiological symptoms of wd it doesn't mean it doesn't fuck up potheads who have to go without smoking for, say, week. Mind your own business.} \\
\midrule
\texttt{Interpersonal Stigma} & \texttt{rehab,\textbf{\textcolor{blue}{sister, family, dad, grandmother, parents, mother, father,}} drugs, mom} & \texttt{\textbf{sadness}} & \texttt{\{`i', `know'\}: 381, \{`i', `have'\}: 260, \{`i', `want'\}: 256} & \texttt{I wish my sister would just go to prison and leave my family alone. About 10 years ago my sister got into a bad wreck. She was in a coma for a week and now has traumatic brain injury.} \\
\midrule
\texttt{Moral Judgments on Addiction} & \texttt{homelessness,\textbf{\textcolor{blue}{homeless, neighbor, neighbors, neighbour, junkies,}} neighborhood, drugs, \textbf{\textcolor{lightgreen}{heroin,}} \textbf{\textcolor{blue}{cops}}} & \texttt{\textbf{annoyance}} & \texttt{\{`i', `see'\}: 23, \{`i', `know'\}: 21, \textbf{\textcolor{red}{\{`i', `hate'\}: 19}}} & \texttt{This is completely ignoring the fact that drugs are the reason they are homeless in the first place. Some of the other comments were saying that they do drugs so why should they judge a homeless person doing drugs. This kind of justification seems insane to me. Just because you are ruining your life, doesn't mean that you should advocate for other people to ruin their lives. And I don't even want to get into the hundreds of drug subreddits like r/heroin, r/meth, and r/crack where people are posting about and bragging about their dangerous drug addictions.} \\
\midrule
\texttt{Moral Judgements and Amphetamine Use} & \texttt{\textbf{\textcolor{lightgreen}{adderall, amphetamine, amphetamines,}} adhd, \textbf{\textcolor{lightgreen}{stimulant, prescriptions, prescription,}} drugs, medication, \textbf{\textcolor{purple}{prescribed}}} & \texttt{neutral} & \texttt{\{`i', `have'\}: 5, \{`i', `had'\}: 4, \textbf{\textcolor{red}{\{`i', `hate'\}: 4}}} & \texttt{I live in a college town and adderall/vyvanse use is insane. Some use it to study, some use it to party and some use it to game for days. All these people eventually can't operate without the pills. It leads to serious rage issues and mood swings. My roommate spends around \$300/month on someone else's adderall. Here are some facts- he will exhaust you with hours of pointless stories and ramblings then get mad when you don't listen. He literally can't shut the hell up. Just like a tweaker.} \\
\midrule
\texttt{Drug Use Consequences} & \texttt{\textbf{\textcolor{lightgreen}{vicodin,}} \textbf{\textcolor{purple}{smoked, smoking,}} toxic, \textbf{\textcolor{purple}{camping, run, thinking,}} scared, needle, \textbf{\textcolor{lightgreen}{crystal}}} & \texttt{neutral} & \texttt{\textbf{\textcolor{red}{\{`i', `wanted'\}: 6},} \{`i', `know'\}: 5, \textbf{\textcolor{red}{\{`it', `feels'\}: 5}}} & \texttt{Shot of meth feels like you've finally crossed that line you swore you'd never cross. You know the one--it looked impossibly far away back when you were naive enough to promise yourself you'd always stick to smoking. When you truly believed you would never allow yourself to become one of those needle freak losers.} \\
\bottomrule
\end{tabular}}
\caption{Summary of topics from direct stigmatizing posts. Interpersonal entities in \textcolor{blue}{blue}, substances in \textcolor{lightgreen}{green}, and actions in \textcolor{purple}{purple}. }
\label{tab:summary1}
\end{table*}

\begin{table*}[t]
\centering
\small
\resizebox{\textwidth}{!}{
\begin{tabular}{>{\raggedright\arraybackslash}p{2cm}>{\raggedright\arraybackslash}p{3.2cm}>{\raggedright\arraybackslash}p{2cm}>{\raggedright\arraybackslash}p{3cm}>{\raggedright\arraybackslash}p{8.5cm}}
\toprule
\textbf{\texttt{Name}} & \textbf{\texttt{Representation}} & \textbf{\texttt{Top Emotion}} & \textbf{\texttt{Top Verb-Subject Pairs}} & \textbf{\texttt{Example}} \\
\midrule
{\texttt{Sobriety \& Family Struggles}} & {\texttt{\textbf{\textcolor{purple}{depressed}}, depression, alcoholic, \textbf{\textcolor{purple}{sober}}, stay, addiction, parents, drinking, \textbf{\textcolor{purple}{quit}}, \textbf{\textcolor{blue}{mother}}}} & {\texttt{\textbf{sadness}}} & \texttt{\{`i', `have'\}: 410, \{`i', `m'\}: 360, \{`i', `want'\}: 345} & \texttt{I've been trying to come out of my isolation, they don't really care, and would rather keep my home and safe. so they screamed at me because I stayed out with my friends too late. I do not have that freedom anymore. I felt like I wanted to stay with my friends until I got comfortable. This was the first time I had hung out with them in a month, and I wasn't even enjoying it. I was uncomfortable. I tried weed, got even more uncomfortable. I can almost never turn down drugs. I am such a pathetic fucking junky.} \\
\midrule
{\texttt{Prescription Medication}} & \texttt{adderall, \textbf{\textcolor{lightgreen}{medications}}, \textbf{\textcolor{lightgreen}{prescription}}, adhd, \textbf{\textcolor{lightgreen}{medication}}, \textbf{\textcolor{lightgreen}{opiate}}, \textbf{\textcolor{purple}{prescribed}}, meds, pharmacy, \textbf{\textcolor{lightgreen}{xanax}}} & {\texttt{\textbf{disappointment}}} & \texttt{\{`i', `have'\}: 78, \textbf{\textcolor{red}{\{`i', `feel'\}: 74}}, \{`i', `know'\}: 46} & \texttt{I apologize if this doesn't make sense. I'm not very good at explaining things. I'm sure a lot of people will just judge me for being a whiny addict and say ``well don't do drugs and you wouldn't even be in this situation, duh''. I get it, most people think that all junkies should be ``thrown on an island to die'' and the world would be a much better place.} \\
\midrule
{\texttt{Overdose Death \& Suicide Ideation}} & \texttt{\textbf{\textcolor{purple}{overdosed}}, \textbf{\textcolor{lightgreen}{xanax}}, \textbf{\textcolor{lightgreen}{fentanyl}}, \textbf{\textcolor{lightgreen}{dilaudid}}, \textbf{\textcolor{lightgreen}{acetaminophen}}, 600mg, 30mg, \textbf{\textcolor{lightgreen}{tramadol}}, \textbf{\textcolor{lightgreen}{prozac}}, \textbf{\textcolor{lightgreen}{clonazepam}}} & {\texttt{\textbf{desire}}} & \texttt{\{`i', `want'\}: 21, \{`i', `m'\}: 15, \{`i', `know'\}: 9} & \texttt{It didn't work, I'm not dead. I looked up what would happen if I took a shit ton of vyvanse and apparently seizures and heart failure are likely. shitty death but I needed to organize my stuff so it's easier to move or get rid of. so I took all the ones I had in the bottle. I spent literally the last couple hours writing suicide notes for nothing.} \\
\midrule
{\texttt{Struggles with Intimate Partners}} & \texttt{youll, bye, alcoholic, \textbf{\textcolor{purple}{leave}}, \textbf{\textcolor{purple}{escaping}}, \textbf{\textcolor{purple} {whisper}}, \textbf{\textcolor{blue}{soul}}, leaving, \textbf{\textcolor{lightgreen}{pot}}, lifennim} & {\texttt{\textbf{sadness}}} & \texttt{\{`i', `want'\}: 11, \{`i', `m'\}: 10, \{`i', `m'\}: 9\}} & \texttt{Just an out of the blue rant from a worthless junkie.... don't bother. oh god I miss you so much. we know each other inside and out and have been through it all. I never thought you'd take me back ever from all the horrible shit I've done then to my surprise you took my back a second time even tho I ran away for months on end with no word or attempt of communication, getting high and drunk 24/7 and randomly showed up back home at 3 in the morning just to leave two days later and repeat my actions. then you moved a whole other province away to get back with me just for me to turn back to drugs and lose my job then you left for the final time.} \\
\bottomrule
\end{tabular}}
\caption{Summary of topics from self-stigmatizing posts. Interpersonal entities in \textcolor{blue}{blue}, substances in \textcolor{lightgreen}{green}, and actions in \textcolor{purple}{purple}.}
\label{tab:summary2}
\end{table*}

\section{Human Evaluation}\label{app: human eval}
We provided the following instructions to guide the evaluation of de-stigmatized texts. These guidelines were shared with our reviewers together with the generated texts from the six systems and forms for ranking the models. Each reviewer judged 20 to 30 posts independently. 

\textbf{Instructions:}
Please read the original post and the generated posts from each of the six systems carefully. For each of the following questions, select the system that best represents your evaluation. Use the space provided for any additional comments you may have.

\subsection*{Overall Quality:}
Assess the overall quality of the generated text with respect to the following measures in addition to de-stigmatization and faithfulness:
\begin{itemize}
  \item \textbf{Naturalness:} The degree to which the output is likely to be used/chosen by a native speaker in the given context/situation.
  \item \textbf{Cohesion:} The text should be a well-organized and coherent body of information, not just a dump of related information. Specifically, the sentences should be connected to one another, maintaining good information flow and logic.
  \item \textbf{Appropriateness:} The degree to which the output as a whole is appropriate in the given context/situation. E.g., ``does the text appropriately consider the parents' emotional state in the given scenario?''
  \item \textbf{Human-Likeness:} The degree to which the output could have been produced by a human.
\end{itemize}
Which system generated the text that with the best overall quality (content, form, de-stigma)?

\subsection*{De-stigmatization:}
Which system's generated post is the most de-stigmatized?

\subsubsection*{Effectively De-stigmatized:}
Which system's post has been the most effectively de-stigmatized, removing negative or harmful stereotypes? Remember stigma is defined as the co-occurring processes of labeling, stereotyping, separation, status loss, and discrimination. In the context of SUD, it can look something like this:
\begin{description}[noitemsep,topsep=0pt]
  \item[Labeling:] involves identifying individuals as different based on certain characteristics—in this case, their substance use. For those with SUD, labels such as ``addict'' or ``alcoholic'' can be affixed.
  \item[Stereotyping:] involves ascribing a fixed set of beliefs or characteristics to individuals based solely on their disorder.
  \item[Separation/Status Loss:] the social distancing of a group perceived as different or undesirable. This separation is partly due to the fear and misunderstanding surrounding the disorder.
  \item[Discrimination:] Discrimination can be both formal and informal, impacting various aspects of life, including employment and social interactions.
\end{description}
\subsection*{Faithfulness:}
Evaluate whether the posts generated by each system contain all the required information from the original post without unnecessary details. Which system has the most faithful result?
\subsection*{General Feedback:}
Please provide any general feedback or additional comments regarding your evaluation of the texts.

\end{document}